\documentclass[a4paper,twoside]{article}

\usepackage{epsfig}
\usepackage{subcaption}
\usepackage{calc}
\usepackage{amssymb}
\usepackage{amstext}
\usepackage{amsmath}
\usepackage{amsthm}
\usepackage{multicol}
\usepackage{multirow}
\usepackage{pslatex}
\usepackage{apalike}
\usepackage{adjustbox}
\usepackage{booktabs}
\usepackage{placeins}
\usepackage{graphicx}
\usepackage[inkscape]{svg}
\graphicspath{{./images/}}
\usepackage[ruled,vlined]{algorithm2e}
\usepackage[bottom]{footmisc}
\usepackage{SCITEPRESS}     

\begin{document}

\title{Self-Supervised Iterative Refinement for Anomaly Detection in Industrial Quality Control}

\author{\authorname{Muhammad Aqeel\sup{1}\orcidAuthor{0009-0000-5095-605X}, Shakiba Sharifi\sup{1}\orcidAuthor{0009-0008-6309-635X}, Marco Cristani\sup{1}\orcidAuthor{0000-0002-0523-6042} and Francesco Setti\sup{1}\orcidAuthor{0000-0002-0015-5534}}
\affiliation{\sup{1}Dept. of Engineering for Innovation Medicine, University of Verona, Strada le Grazie 15, Verona, Italy}
\email{muhammad.aqeel@univr.it}
}

\keywords{Robust Anomaly Detection, Self-Supervised Learning, Iterative Refinement Process, Industrial Quality Control}

\abstract{This study introduces the Self-Supervised Iterative Refinement Process (IRP), a robust anomaly detection methodology tailored for high-stakes industrial quality control. The IRP leverages self-supervised learning to improve defect detection accuracy by employing a cyclic data refinement strategy that iteratively removes misleading data points, thereby improving model performance and robustness. We validate the effectiveness of the IRP using two benchmark datasets, Kolektor SDD2 (KSDD2) and MVTec-AD, covering a wide range of industrial products and defect types. Our experimental results demonstrate that the IRP consistently outperforms traditional anomaly detection models, particularly in environments with high noise levels. This study highlights the potential of IRP to significantly enhance anomaly detection processes in industrial settings, effectively managing the challenges of sparse and noisy data.}

\onecolumn \maketitle \normalsize \setcounter{footnote}{0} \vfill

\section{\uppercase{Introduction}}
\label{sec:introduction}

Anomaly detection (AD) plays an indispensable role in quality control in a wide range of manufacturing industries, ensuring the integrity of materials such as marble~\cite{vrochidou2022marble}, steel~\cite{Bozic2021COMIND}, and leather~\cite{jawahar2023leather}. Accurately detecting items that are non-compliant with product specifications is crucial for maintaining product standards and consumer satisfaction. However, this task is fraught with challenges, primarily due to the diverse and complex textures of the materials, the rarity of defects, and the significant scarcity of accurately labeled data necessary for effective supervised learning. Anomalies typically occupy only a tiny portion of an image, making their detection difficult for human inspectors and automated systems. This challenge is compounded by the labor intensive and error-prone nature of manually labeled training data, a process that becomes increasingly unsustainable in high-throughput manufacturing settings.

The evolution of anomaly detection in recent years has increasingly leaned towards unsupervised and self-supervised learning paradigms, prized for their ability to operate without extensively labeled datasets. These approaches, while innovative, presuppose the existence of pristine training data, free from anomalies~\cite{ono2020robust,beggel2020robust}. This ideal is seldom met in practical scenarios, where the inclusion of even a few anomalous samples can severely skew the learning process, leading to either too restrictive or overly permissive models. Models trained under these conditions are prone to overfitting, resulting in many false positives during operational deployment. In contrast, including anomalies in the training set can escalate the incidence of false negatives, severely undermining the system's reliability.

\begin{figure*}[ht]
\centering
  \includegraphics[width=\linewidth]{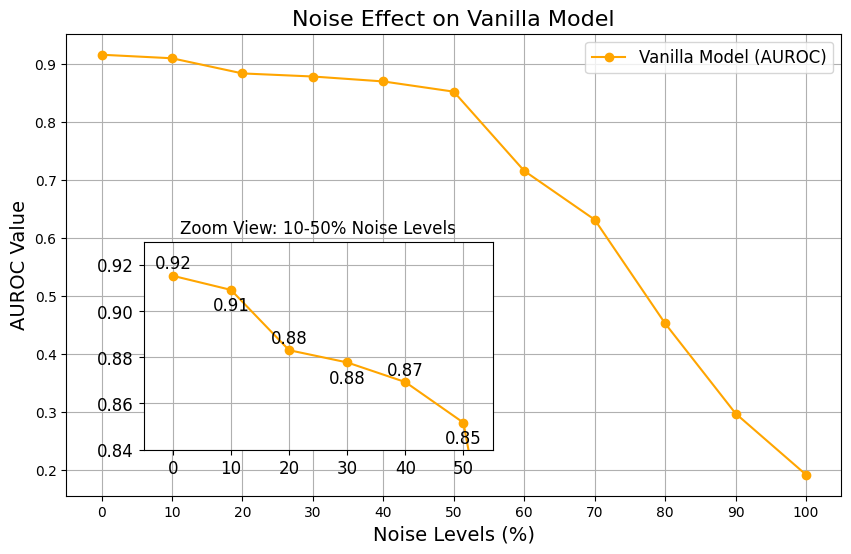}
  \caption{Typical impact of noisy data in training an anomaly detection model.}
\label{fig}
\end{figure*}

To illustrate the severity of this issue, Fig.~\ref{fig} demonstrates a decrease in AUCROC scores as noise increases in the training data. Traditional detection methods show a marked deterioration in performance as data imperfections increase, highlighting their vulnerability. In contrast, our proposed Self-Supervised Iterative Refinement Process (IRP) maintains a higher level of performance, showcasing its robustness and adaptability across a broad range of challenging conditions. We deliberately limit our evaluation to scenarios where the noise does not render the data overly corrupt, ensuring meaningful learning and generalization. This decision reflects our commitment to developing a method that optimally balances performance with practical applicability in realistically noisy environments.

Addressing these complexities requires a robust detection system capable of navigating the intricacies of noisy and incomplete data sets. Current advances in deep learning, such as autoencoders~\cite{zavrtanik2022dsr}, generative adversarial networks (GAN)~\cite{luo2022robust}, and discrete feature space representations~\cite{hu2022robust}, offer promising ways to improve anomaly detection. However, these technologies still struggle with determining the optimal anomaly score function, ensuring sufficient robustness against noise and outliers, and generalizing effectively to novel, unseen data sets.

In response to these ongoing challenges, this paper introduces a methodology named \emph{Self-Supervised Iterative Refinement Process (IRP)} specifically designed for robust anomaly detection. This process combines advanced machine learning techniques with a cyclic data refinement strategy to iteratively enhance the training data set quality, thereby significantly improving the model's performance and robustness. IRP systematically identifies and removes the most misleading data points based on a dynamically adjusted scoring mechanism, thereby refining the model's ability to generalize from normal operational data while minimizing the influence of outliers.

The main contributions of this paper are as follows.
\begin{itemize}
  \item We propose the Self-Supervised Iterative Refinement Process (IRP), a robust training methodology that employs a cyclic refinement strategy to enhance data quality and model accuracy in anomaly detection systematically;
  \item The dynamic threshold adjustment mechanism adapts the exclusion threshold based on a robust statistical measure, ensuring precise and adaptive outlier removal. This feature improves the system’s ability to accurately identify and exclude data points that do not conform to the expected pattern, improving overall detection performance.
  \item We present an experimental validation of our proposed methodology in two challenging public datasets, namely KSDD2 and MVTec-AD, which demonstrate significant improvements over traditional approaches and establish a new standard for robustness in the field.
\end{itemize}

\section{\uppercase{Related Work}}

Detecting and identifying defects is crucial in manufacturing to ensure processes function correctly. Anomaly detection involves identifying issues such as scratches, blemishes, blockages, discoloration, holes, and, in general, any surface or structural irregularity on a manufactured item~\cite{de2020row}. Recent surveys review AD's state of the art from both technological and methodological perspectives~\cite{chen2021surface,bhatt2021image}. 
Traditionally, AD can be formulated as a supervised or unsupervised learning problem. Supervised methods use both anomaly and normal samples during training, while unsupervised methods only use normal samples at training time. While supervised methods achieve better results, the high cost of annotated datasets and advances in generative AI have shifted focus towards unsupervised approaches. Many unsupervised methods rely on image reconstruction using encoder-decoder networks to spot anomalies by comparing original and reconstructed images~\cite{akcay2019ganomaly,defard2021padim,roth2022towards}. These networks struggle to reconstruct anomalous regions accurately, having never encountered them during training, which can result in high false positive rates with variable training data. 
To address this, \cite{zavrtanik2021draem} employs a reconstruction network to restore normal appearances in anomalous images, while another approach introduces textual prompts describing defects as in \cite{capogrosso2024diffusion,girella2024cbmi}. 
Comparing features instead of whole images also helps to reduce false positives~\cite{zavrtanik2022dsr,rudolph2021same}.

Robust anomaly detection (RAD) addresses the problem of mitigating the effect of bad annotations in the training set. It has been widely studied in several fields, using methods ranging from robust statistical techniques~\cite{rousseeuw2018anomaly} to deep learning approaches like autoencoders~\cite{beggel2020robust,zhou2017anomaly} and recurrent neural networks~\cite{su2019robust}. These methods all aim to reduce sensitivity to noise in the labeling process. 
Alternatively, \cite{zhao2019rad} focused on removing noisy labels while using high-performing anomaly detectors. Here, a two-layer online learning framework filters suspicious data and detects anomalies from the remaining data. We argue that training two different models (one for predicting data quality and one for anomaly detection) adds unnecessary complexity. 
More recently, researchers have explored self-supervised learning (SSL) as a promising avenue for anomaly detection. SSL methods often rely on adaptations of deep CNN for enhancing learning capabilities from unlabelled data ~\cite{bovzivc2021end,zhang2022steel,tian2022dcc}, or using domain-adaptation techniques~\cite{zhang2021visual}. In our previous work~\cite{aqeel2024delta}, we demonstrated that SSL could be effective in filtering badly labeled data in the training set. This paper extends our previous work by using an iterative refinement approach, which is based on statistical observations of anomaly score estimation on the training set, to further improve robustness and accuracy in anomaly detection.

\begin{figure*}[htbp]
    \centering
   \includegraphics[width=.9\linewidth]{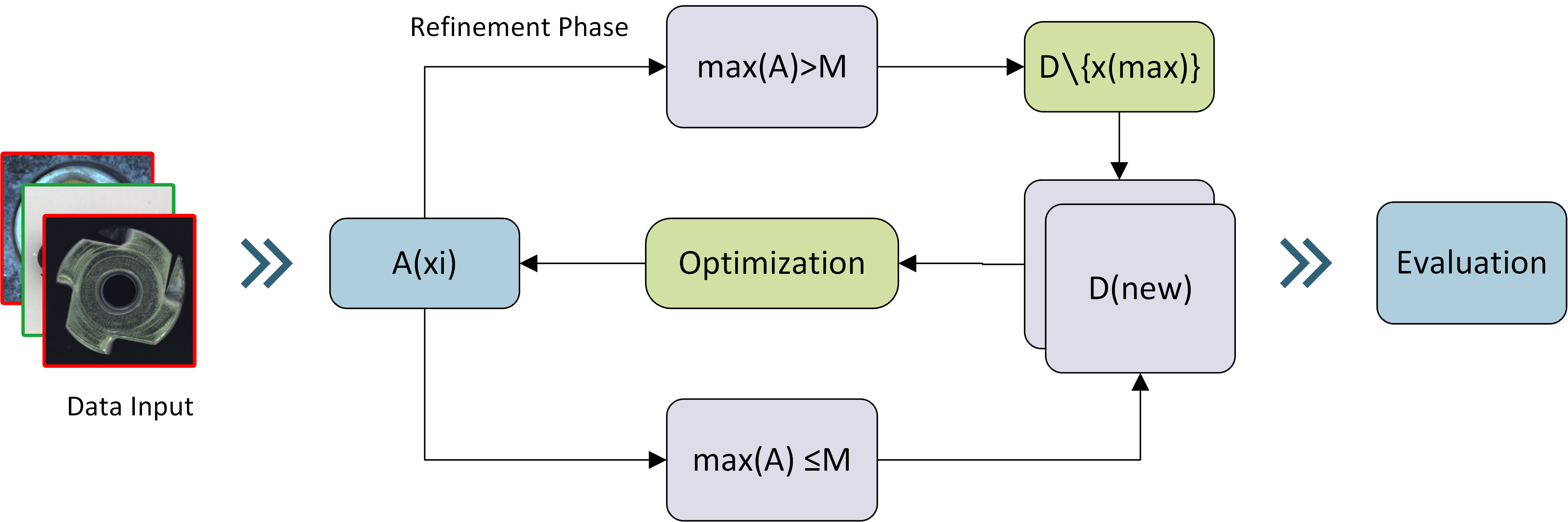}
    \caption{Illustrative schematic of the Iterative Refinement Process, demonstrating the independent cycle of training, validation, retraining, and testing.}
    \label{fig:concept}
\end{figure*}

\section{\uppercase{Iterative Refinement Process}}
The Iterative Refinement Process (IRP) introduces a novel approach to significantly enhance the robustness and accuracy of anomaly detection in an industrial environment by applying advanced probabilistic models to iteratively refine the training dataset. Despite being agnostic with respect to the anomaly detection model, for our preliminary evaluation of the impact of our approach, we used the DifferNet model~\cite{rudolph2021same}, which provided foundational insights into the potential refinements necessary for handling complex anomaly detection in industrial settings. The DifferNet model employs a normalizing flow-based architecture to provide precise density estimations of image features extracted via a convolutional neural network. DifferNet establishes a bijective relationship between the feature space and the latent space, enabling each vector to be uniquely mapped to a likelihood score. This scoring is derived from the model's ability to discern common patterns from uncommon ones, making it especially effective for detecting subtle anomalies often found in defect detection scenarios. Initially, each data point $x_i$, which refers to a transformed image, is evaluated using a probabilistic model tailored to anomaly detection. The model calculates an anomaly likelihood as follows:
\begin{equation}
  A(x_i) = -\log p(x_i; \theta)
\end{equation}
where $p(x_i; \theta)$ is the probability density function of the transformed image $x_i$, with $\theta$ representing the parameters of the model. The framework engages in a dynamic data refinement cycle, where each point is assessed against an adaptive threshold calculated as $\lambda$ times the median of the anomaly scores, $A(x_i)$. If any data point’s anomaly score exceeds this threshold, it is removed to refine the dataset. The refined dataset, denoted \(D_{\text{new}}\), excludes the most anomalous data point, \(x_{\text{max}}\), where \(x_{\text{max}}\) is the point with the maximum anomaly score exceeding the threshold. This process of refinement and retraining on \(D_{\text{new}}\) continues until the dataset achieves optimal stability and performance. 
The steps involved in this refinement process are elaborated in Algorithm~\ref{algo}, and the entire sequence from data input through anomaly detection, refinement, and final evaluation is illustrated in Fig.~\ref{fig:concept}.

\subsection{Probabilistic Anomaly Scoring Mechanism and Dynamic Refinement}
Integral to DifferNet's approach is a robust probabilistic scoring system that enhances the accuracy and robustness of defect detection through dynamic refinement processes. This anomaly detection framework utilizes a convolutional neural network (CNN) as a pretrained feature extractor, denoted as \( f_{ex} \), which is not optimized further post-training. This extractor transforms input data \( x \in X \) into a feature space \( Y \), specifically:
\begin{equation}
    y = f_{ex}(x) \in Y
\end{equation}

These extracted features \( y \) are further processed by a state-of-the-art normalizing flow model \( f_{NF} \), which maps them into a well-defined latent space \( Z \) with a known distribution \( p_Z(z) \). This transformation is crucial as it enables the precise estimation of the data's likelihood based on the latent representation:
\begin{equation}
    z = f_{NF}(y) \quad \text{where} \quad z \sim \mathcal{N}(0, I)
\end{equation}


The anomaly scores are derived directly from the latent space by evaluating the likelihood of the transformed features. A lower likelihood indicates a higher probability of an anomaly, as it reflects the deviation of the features from the model of 'normal' data. These scores are computed by averaging the negative log-likelihoods of multiple transformed versions of an image $x$, which are generated using a set of predefined transformations $T$. The transformations $T$ represent a collection of operations that perturb or modify the original data in controlled ways to simulate variations that may highlight anomalies more clearly. These transformations may include geometric modifications, such as rotations, translations, and scalings, or they may involve noise injections or other perturbations that simulate real-world variations in the data.

The purpose of applying these transformations is twofold. First, it helps with data augmentation by generating multiple alternative versions of the input data, thereby making the anomaly detection process more robust to specific features of the original data. Second, it stabilizes the anomaly scoring by reducing the impact of random noise introduced by any single transformation. By averaging the scores over several transformed instances of the data, the final anomaly score becomes more reliable, reflecting the underlying anomaly rather than spurious noise.

For each transformation $T_i \in T$, the anomaly score is calculated by applying $T_i$ to the input data $x$, passing it through the feature extractor $f_{ex}$, and then mapping the resulting features to the latent space using the normalizing flow model $f_{NF}$. The final score is the expected value of the negative log-likelihoods of these transformed instances, as shown in Equation~\ref{eq:scoring}. This approach ensures that the anomaly score reflects the data's likelihood in the transformed space, improving the robustness of the anomaly detection process by incorporating multiple views of the data.
\begin{equation}
    \tau(x) = \mathbb{E}_{T_i \in T}[-\log p_Z(f_{NF}(f_{ex}(T_i(x))))] \label{eq:scoring}
\end{equation}

Our significant contribution is manifest in the dynamic adjustment of the assessment threshold \( \theta \), which is iteratively recalibrated in response to evolving data characteristics and directly influenced by the statistical nature of the anomaly scores. We chose the median as a central measure for thresholding due to its robustness against outliers and skewed data distributions. The median offers a more stable and representative central tendency, particularly in datasets where the majority of values are concentrated around a certain range with a few extreme outliers. This characteristic makes the median especially suitable for determining thresholds in anomaly detection, where the presence of anomalies can distort the distribution of scores. The threshold defined as:
\begin{equation}
    \text{Threshold} = \lambda \times M \label{eq:threshold}
\end{equation}
where $\lambda$ is the scaling factor and $M$ is the median of the anomaly scores. This ensures that data points with anomaly scores exceeding the threshold are considered outliers and are removed from the dataset.
\begin{equation}
    D_{\text{new}} = D_{\text{old}} \setminus \{x \mid \tau(x) > \text{Threshold}\} \label{eq:refinement}
\end{equation}

This iterative refinement process not only enhances the quality and performance of the model but also stabilizes it at an optimal level. Moreover, this method is designed not only to detect but also localize anomalies within images, leveraging the gradient of the loss function to clearly demarcate regions contributing to anomalies, thus providing a robust mechanism for practical deployment in industrial quality control systems.

\begin{algorithm}[t]
\caption{Iterative Refinement and Retraining}
\DontPrintSemicolon 

{Dataset of images $D = \{x_1, x_2, \ldots, x_n\}$}
Pre-train model on $D$ for $N$ epochs\;
Compute anomaly scores $A = \{a_1, a_2, \ldots, a_n\}$ where $a_i = -\log p(x_i; \theta)$\;
Compute median of scores $M = \text{median}(A)$\;
\If{$\max(A) > \lambda \times M$}{
    $x_{\text{max}} \gets x_i$ where $a_i = \max(A)$\;
    Delete $x_{\text{max}}$ from $D$\;
    $D_{\text{new}} \gets D \setminus \{x_{\text{max}}\}$\;
}
\Else{
    $D_{\text{new}} \gets D$\;
}
Retrain model on $D_{\text{new}}$\;
\Repeat{Convergence}{
    Repeat steps 3 to 9\;
}
\label{algo}
\end{algorithm}

\section{\uppercase{Experimental Results}}
This section presents the experimental results obtained from applying the Iterative Refinement Process (IRP) to anomaly detection. The experiments were designed to evaluate the effectiveness of the IRP in improving detection accuracy and robustness under various conditions.

\subsection{Datasets}
We assess the performance of our approach using two publicly available datasets, which are extensively used to assess the robustness and effectiveness of defect detection models:

\textbf{Kolektor SDD2 (KSDD2):} The KSDD2 dataset \cite{Bozic2021COMIND} consists of RGB images showcasing defective production items, meticulously sourced and annotated by Kolektor Group d.o.o. The defects vary widely in size, shape, and color, ranging from minor scratches and spots to significant surface faults. For uniform evaluation, all images are center-cropped and resized to dimensions of 448$\times$448 pixels. The training set comprises 2085 normal and 246 positive samples, while the testing set includes 894 negative and 110 positive samples.

\textbf{MVTec-AD Dataset:} The MVTec-AD dataset \cite{bergmann2019mvtec} is utilized to demonstrate the effectiveness of our proposed method across a variety of real-world industrial products. The MVTec-AD dataset contains 15 different types of industrial products, encompassing over 70 different types of defects, each labelled with both defect types and segmentation masks. To address the original lack of defective images in the training set, we resampled half of the defect images to include in our training data. The revised training set contains 889 normal and 1345 defective images, while the testing set comprises 1210 normal and 724 defective images.

\begin{figure*}[ht]
    \centering
    \subfloat[KSDD2]{\includegraphics[width=0.49\linewidth]{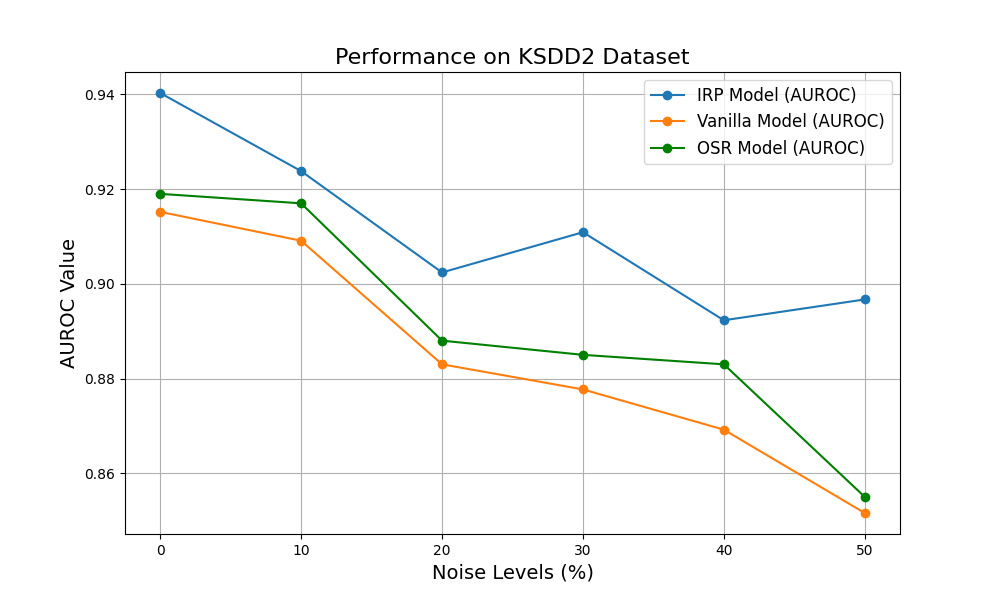}}\label{ksdd2}
    \hfill
    \subfloat[MVTec-AD]{\includegraphics[width=0.49\linewidth]{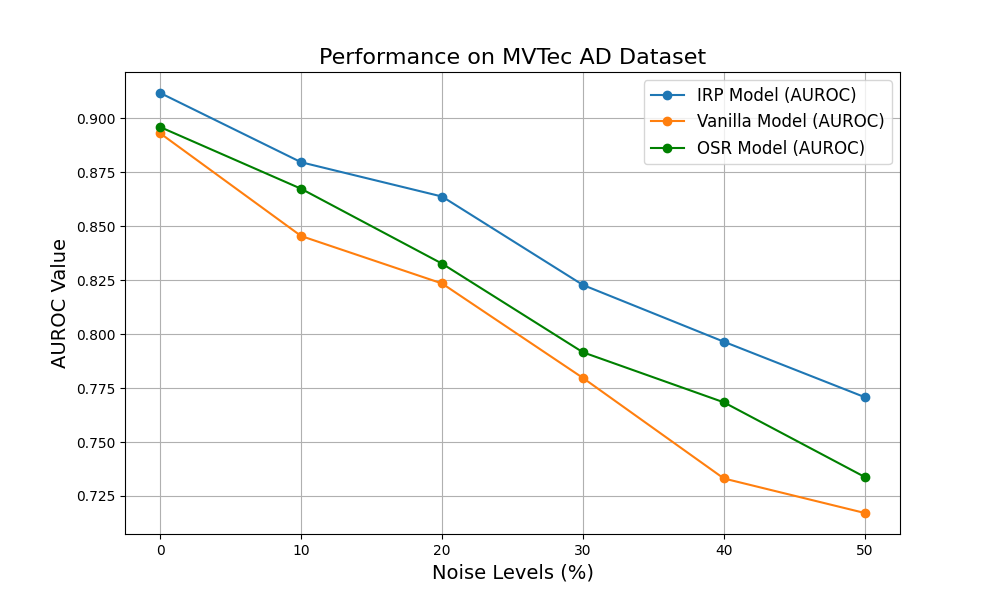}}\label{mvtec}
    \caption{Figure (a) displays the performance trends of various models on the KSDD2 dataset, while figure (b) shows results for the MVTec-AD dataset, across different noise levels. The graphs illustrate the models' performance before and after applying the IPR model: the blue line represents the IPR model, the green line denotes the OSR model, and the yellow line illustrates the performance of the vanilla model.}
    \label{performance}
\end{figure*}

\subsection{Implementation Details} 
Experiments were conducted using PyTorch framework. We resized images to \(448 \times 448\) pixels and applied standardized preprocessing techniques, including optional rotations and normalization.
The network architecture features three scales of input resolution, 8 coupling blocks within the normalizing flow model, and fully connected layers in the scale-translation networks each containing 2048 neurons, with no dropout our proposed method's effectiveness across variousapplied. Training parameters are set with a learning rate of \(2 \times 10^{-4}\), and the model undergoes 400 epochs with batch size 96. 

We compared our proposed methodology with existing models including DifferNet~\cite{rudolph2021same} and One Shot Removal (OSR)~\cite{aqeel2024delta}. DifferNet leverages a multi-scale feature extraction process to assign meaningful likelihoods to images, thereby facilitating defect localization. DifferNet represents also the model for feature extraction used in our self supervised procedure. OSR enhances the robustness of surface defect detection models through a novel training pipeline, which involves initial training, anomalous sample removal, and model fine-tuning.

The performance of our model was tested under varying noise levels, from 0\% to 50\%. These noise levels simulate the inclusion of bad samples typically encountered in unsupervised learning scenarios. We measured the model's performance using the AUROC score across these noise levels, providing a reliable indicator of its ability to distinguish between normal and anomalous data. Each experiment was repeated three times to ensure consistency and to assess the impact of noise on detection accuracy.

\subsection{Performance on KSDD2 Across Varying Noise Levels}
This section presents the evaluation results of the IRP on the KSDD2 dataset. The analysis explores the performance of the IRP model under different noise levels, highlighting its defect detection capabilities in challenging environments. As shown in Fig.~\ref{performance}, the IRP model is compared against two conventional approaches—the traditional vanilla model (DifferNet) and the One Shot Removal (OSR) model proposed in~\cite{aqeel2024delta}. Across all tested noise levels, the IRP model consistently outperforms the competitors, demonstrating remarkable resilience by maintaining superior AUROC values, as systematically documented in Table~\ref{tab:performance_medians}.


The IRP model excels particularly at higher noise levels. For example, at 50\% noise, the IRP model achieves a significantly higher AUROC of 0.8967 on the KSDD2 dataset, while the vanilla and OSR models show notable declines. This robustness stems from the IRP's advanced noise-cancellation algorithms, adaptive thresholding, and self-supervised feature extraction, which enable effective handling of heavily corrupted data.

Moreover, the statistical analysis accompanying our findings reinforces the IRP model’s superiority. The standard deviation values reported in Table~\ref{tab:performance_medians} indicate lower variability in performance across different experimental runs, underscoring the model's reliability and predictability. Such attributes are indispensable in industrial applications, where the cost of false negatives can be prohibitive, and the ability to detect anomalies reliably is paramount. The comprehensive data presented in the table not only showcases the IRP model’s adaptability and robust performance across diverse and challenging scenarios but also highlights its viability as a potent tool for maintaining stringent quality control standards in manufacturing processes. These findings suggest that the IRP, through its self-supervised learning approach, can be effectively deployed in industrial settings to enhance defect detection where traditional models might falter due to high variability in defect rates and challenging noise conditions.

\begin{table*}[ht]
\centering
\caption{AUROC scores of the IRP model at different noise levels for MVTec-AD and KSDD2 datasets.}
\label{tab:performance_medians}
\begin{tabular}{@{}clccccccc@{}}
\toprule
\multirow{2}{*}{\textbf{Category}} & \multirow{2}{*}{\textbf{Entity}} & \multicolumn{6}{c}{\textbf{Noise Level}} \\ 
\cmidrule(lr){3-8}
& & \textbf{0\%} & \textbf{10\%} & \textbf{20\%} & \textbf{30\%} & \textbf{40\%} & \textbf{50\%} \\ 
\midrule
\multirow{10}{*}{\textbf{Classes}} 
& \textbf{Bottle} & 0.9386 & 0.9495 & 0.9250 & 0.9200 & 0.9295 & 0.9340 \\
& \textbf{Capsule} & 0.8592 & 0.8498 & 0.8288 & 0.8102 & 0.8012 & 0.7872 \\
& \textbf{Cable} & 0.9288 & 0.8368 & 0.8560 & 0.6756 & 0.5716 & 0.4604 \\
& \textbf{Carpet} & 0.8386 & 0.7727 & 0.7222 & 0.6354 & 0.5905 & 0.6206 \\
& \textbf{Leather} & 0.9722 & 0.9388 & 0.9348 & 0.9266 & 0.9157 & 0.8897 \\
& \textbf{MetalNut} & 0.8880 & 0.8688 & 0.8480 & 0.8464 & 0.7792 & 0.8096 \\
& \textbf{Pill} & 0.8888 & 0.8580 & 0.8558 & 0.8016 & 0.8096 & 0.7816 \\
& \textbf{Screw} & 0.8938 & 0.8680 & 0.8524 & 0.8234 & 0.7988 & 0.7436 \\
& \textbf{Tile} & 0.9723 & 0.9674 & 0.9649 & 0.9656 & 0.9598 & 0.9481 \\
& \textbf{Zipper} & 0.9378 & 0.8880 & 0.8504 & 0.8232 & 0.8096 & 0.7340 \\
\midrule
\multirow{4}{*}{\textbf{Models}} 
& \textbf{Vanilla} & 0.8931 & 0.8455 & 0.8235 & 0.7797 & 0.7331 & 0.7171 \\
& \textbf{OSR} & 0.8960 & 0.8674 & 0.8327 & 0.7916 & 0.7684 & 0.7338 \\
& \textbf{IRP on MVTec-AD} & 0.9118 & 0.8797 & 0.8638 & 0.8228 & 0.7965 & 0.7708 \\
& \textbf{IRP on KSDD2} & 0.9403 & 0.9238 & 0.9024 & 0.9109 & 0.8923 & 0.8967 \\
\midrule
& \textbf{Standard Deviation} & ±0.0246 & ±0.0149 & ±0.0086 & ±0.0518 & ±0.0777 & ±0.1099 \\ 
\bottomrule
\end{tabular}%
\end{table*}

\subsection{Performance on MVTec Across Varying Noise Levels}
The robustness and adaptability of IRP have been rigorously evaluated across various product classes within the MVTec-AD dataset, including Bottle, Cable, Capsule, Carpet, Leather, MetalNut, Pill, Screw, Tile, and Zipper. Each class, with its distinct defect types and unique challenges, served as an ideal testbed for the IRP. We systematically varied the noise levels in the dataset from 0\% to 50\% to meticulously assess the IRP's capabilities under progressively challenging conditions, using the Area Under the Curve of the Receiver Operating Characteristic (AUROC) to quantify defect detection accuracy across different thresholds.

\begin{figure*}[tbp]
\centering
\subfloat[Bottle]{\includegraphics[height=4cm, width=0.49\textwidth]{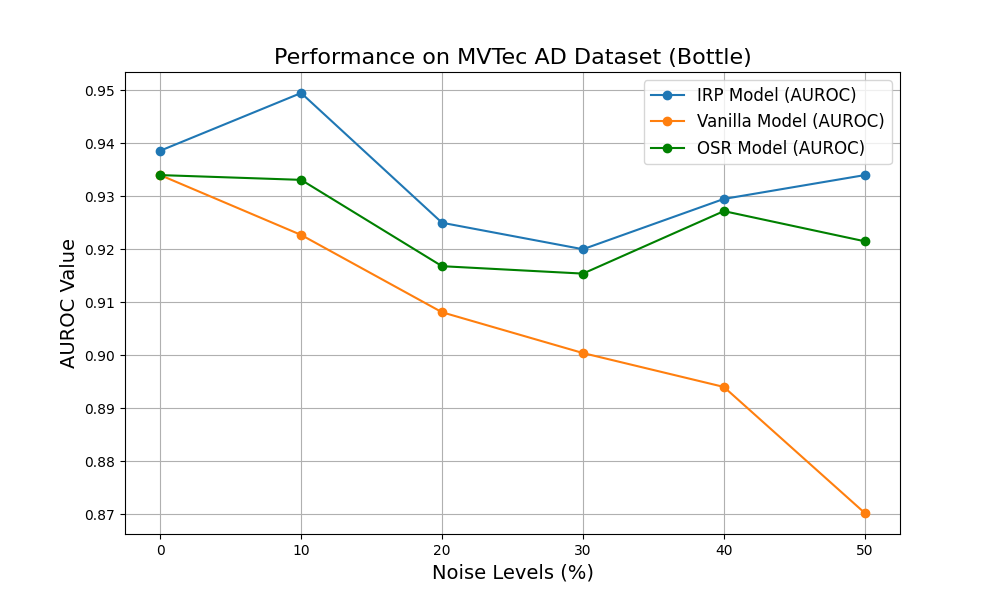}\label{fig:bottle}}
\hfill
\subfloat[Cable]{\includegraphics[height=4cm, width=0.49\textwidth]{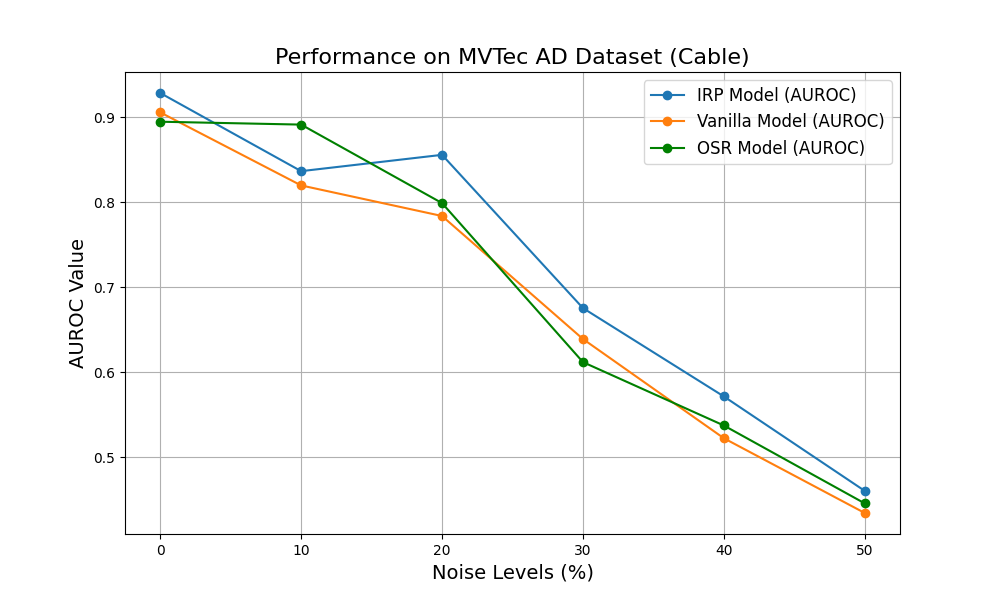}\label{fig:cable}}
\hfill 
\subfloat[Capsule]{\includegraphics[height=4cm, width=0.49\textwidth]{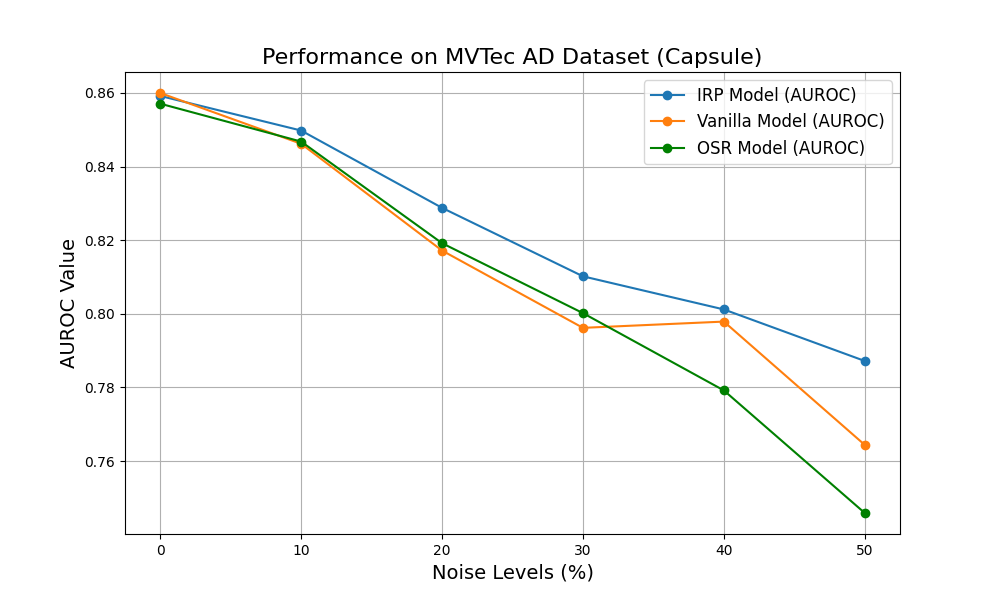}\label{fig:capsule}}
\hfill
\subfloat[Carpet]{\includegraphics[height=4cm, width=0.49\textwidth]{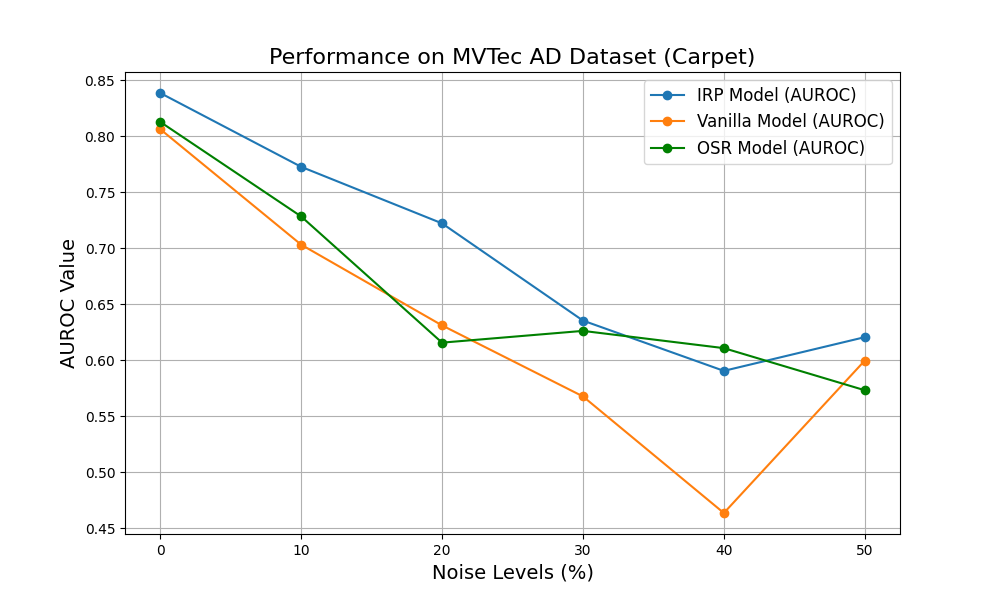}\label{fig:carpet}}
\hfill
\subfloat[Leather]{\includegraphics[height=4cm, width=0.49\textwidth]{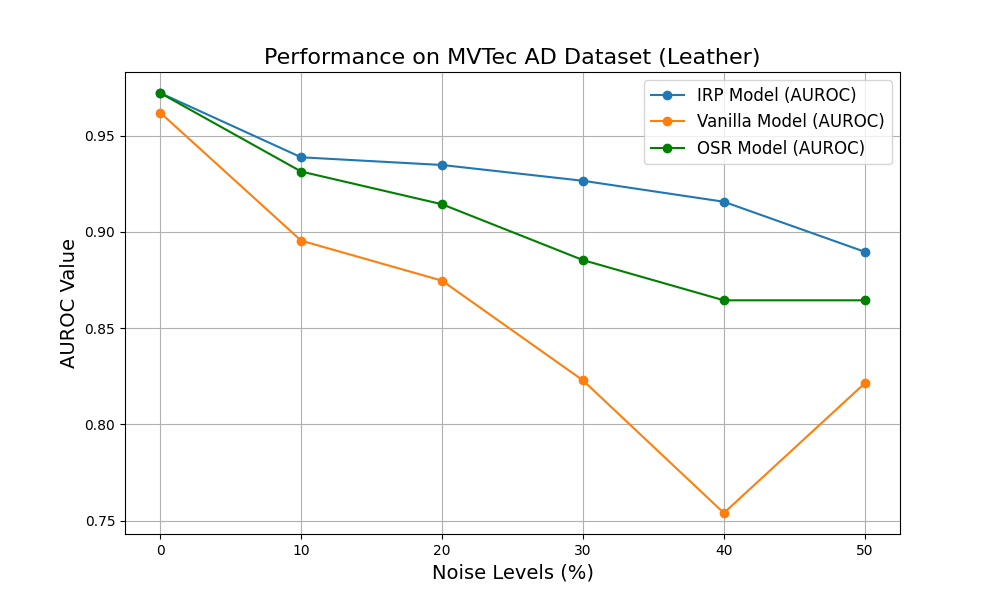}\label{fig:leather}}
\hfill
\subfloat[MetalNut]{\includegraphics[height=4cm, width=0.49\textwidth]{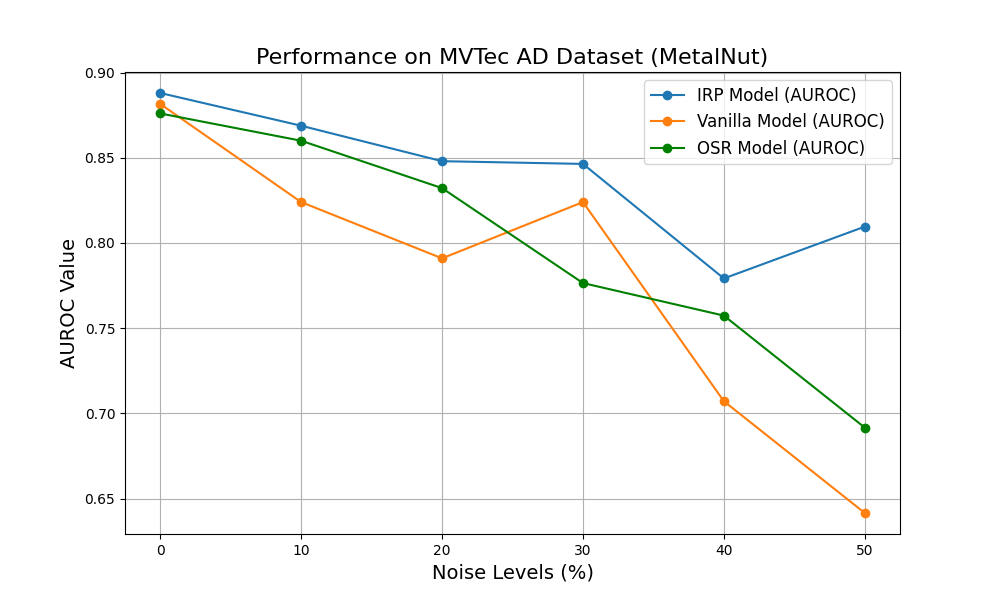}\label{fig:metalnut}}
\hfill
\subfloat[Pill]{\includegraphics[height=4cm, width=0.49\textwidth]{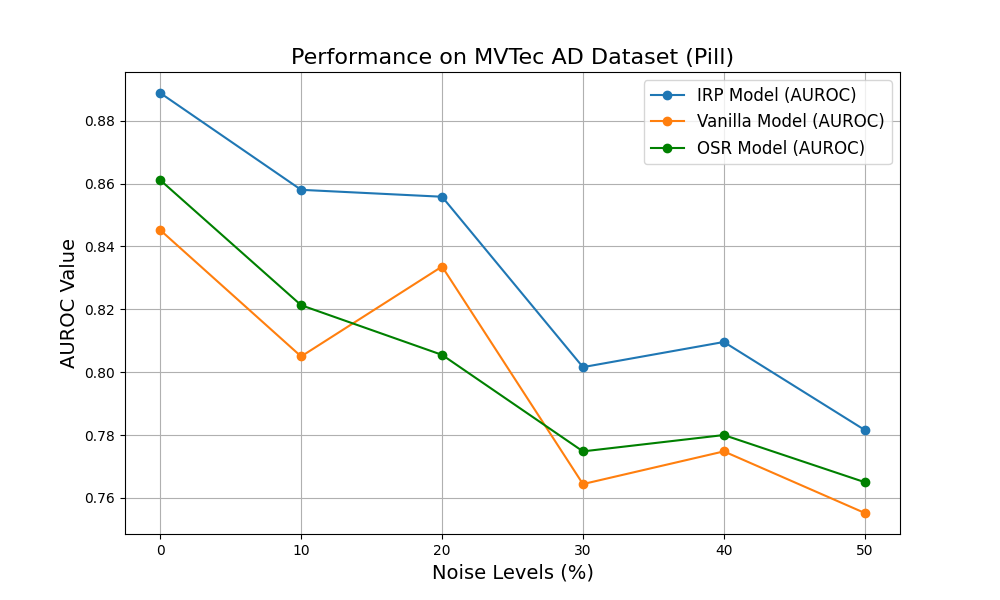}\label{fig:pill}}
\hfill
\subfloat[Screw]{\includegraphics[height=4cm, width=0.49\textwidth]{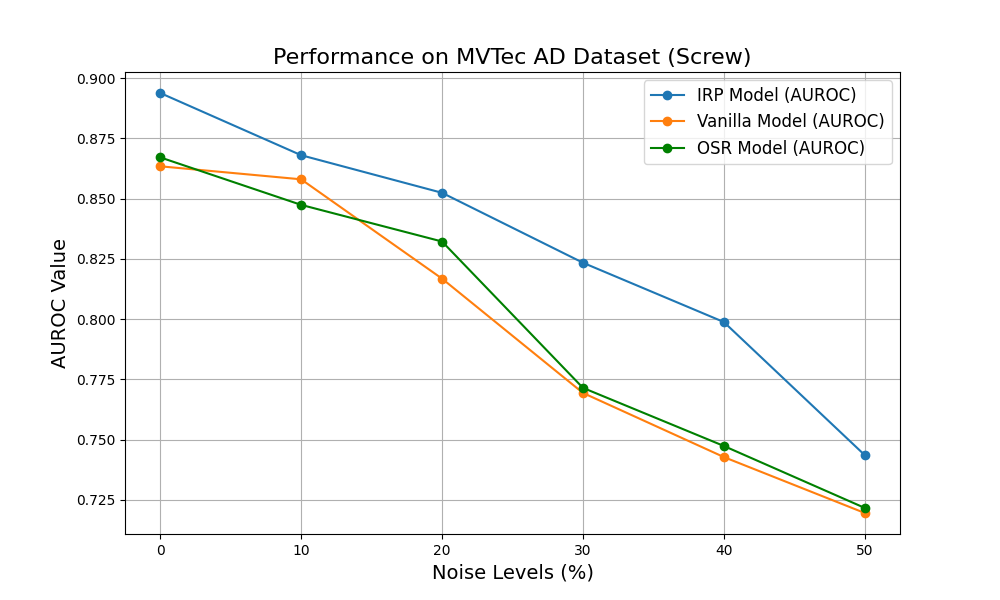}\label{fig:screw}}
\hfill
\subfloat[Tile]{\includegraphics[height=4cm, width=0.49\textwidth]{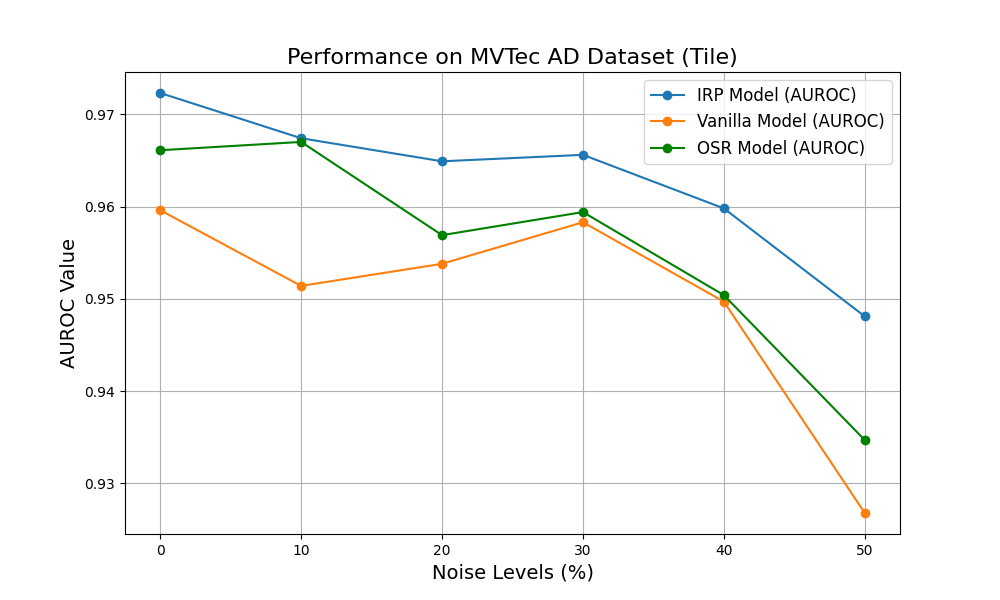}\label{fig:tile}}
\hfill
\subfloat[Zipper]{\includegraphics[height=4cm, width=0.49\textwidth]{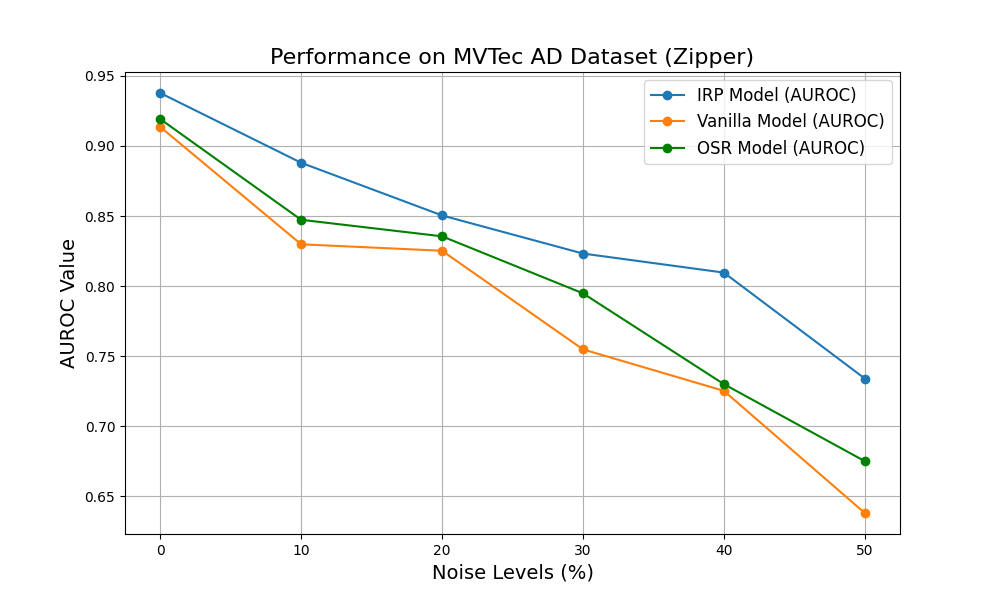}\label{fig:zipper}}
\caption{AUROC performance of the IRP across various classes of the MVTec-AD dataset. Each sub-figure demonstrates the defect detection efficacy for a different class, illustrating the robustness and adaptability of the IRP in handling diverse defect characteristics.}
\label{fig:mvtec_performance}
\end{figure*}

As illustrated in Fig.~\ref{fig:mvtec_performance}, the IRP model consistently achieves high AUROC values across all classes, demonstrating its robust ability to manage noise effectively. Notably, in classes with complex and variable defect characteristics such as Screw and Cable, the IRP exhibits exceptional resilience, showcasing its advanced anomaly detection capabilities. Conversely, in classes like Capsule and Pill, where defects are more subtle, the IRP effectively distinguished between normal variations and actual defects, underlining its sophisticated feature extraction capabilities.

\begin{table*}[tbp]
    \small
    \centering
    \caption{Images deleted at various noise levels for KSDD2 and MVTec-AD datasets, segmented by product categories.}
    \label{tab:performance_metrics}
    \resizebox{\textwidth}{!}{%
    \begin{tabular}{@{}ccccccccccccccc@{}}
    \toprule
    \multirow{3}{*}{\textbf{Noise Level (\%)}} & \multicolumn{2}{c}{\textbf{KSDD2}} & \multicolumn{12}{c}{\textbf{MvTec}} \\
    \cmidrule(lr){2-3} \cmidrule(lr){4-15}
    & \multicolumn{2}{c}{\textbf{}} & \multicolumn{2}{c}{\textbf{Cable}} & \multicolumn{2}{c}{\textbf{Capsule}} & \multicolumn{2}{c}{\textbf{Metalnut}} & \multicolumn{2}{c}{\textbf{Pill}} & \multicolumn{2}{c}{\textbf{Screw}} & \multicolumn{2}{c}{\textbf{Zipper}} \\
    \cmidrule(lr){2-3} \cmidrule(lr){4-5} \cmidrule(lr){6-7} \cmidrule(lr){8-9} \cmidrule(lr){10-11} \cmidrule(lr){12-13} \cmidrule(lr){14-15}
    & \textbf{Good} & \textbf{Bad} & \textbf{Good} & \textbf{Bad} & \textbf{Good} & \textbf{Bad} & \textbf{Good} & \textbf{Bad} & \textbf{Good} & \textbf{Bad} & \textbf{Good} & \textbf{Bad} & \textbf{Good} & \textbf{Bad} \\
    \midrule
    0\%  & 11 & 0 & 6 & 0 & 16 & 0 & 14 & 0 & 17 & 0 & 15 & 0 & 19 & 0 \\
    10\% & 3 & 5 & 5 & 7 & 11 & 6 & 6 & 8 & 9 & 8 & 12 & 5 & 10 & 5 \\
    20\% & 3 & 7 & 5 & 9 & 5 & 12 & 5 & 11 & 5 & 11 & 6 & 9 & 8 & 9 \\
    30\% & 2 & 8 & 4 & 11 & 4 & 14 & 3 & 11 & 4 & 13 & 7 & 11 & 5 & 11 \\
    40\% & 4 & 12 & 3 & 12 & 2 & 13 & 2 & 9 & 5 & 12 & 6 & 10 & 6 & 14 \\
    50\% & 4 & 13 & 4 & 12 & 5 & 13 & 1 & 9 & 3 & 11 & 2 & 12 & 5 & 13 \\
    \bottomrule
    \end{tabular}}\label{tab2}
\end{table*}

This uniform and robust performance across varied classes and noise levels highlights the IRP model’s adaptability and effectiveness in industrial applications, especially in environments with diverse and unpredictable defect rates. The IRP's ability to maintain high accuracy at elevated noise levels underscores its potential as a dependable tool for quality control in manufacturing environments. Furthermore, Fig.~\ref{performance} provides an overarching view of the IRP’s performance across the MVTec-AD dataset. This visual representation emphasizes the IRP’s superior performance relative to the vanilla and OSR models, mainly as it consistently improves defect handling under increasingly challenging conditions. This enhancement affirms the IRP's role as a robust and reliable quality control instrument in complex manufacturing landscapes, where defect rates fluctuate significantly. Table~\ref{tab:performance_medians} complements these insights by detailing the AUROC scores at various noise levels for the MVTec-AD and KSDD2 datasets, including measures of error magnitude. This detailed presentation not only confirms the performance stability of the IRP across different settings but also highlights the model’s reliability, evidenced by the low standard deviation in its performance metrics.

\subsection{Quantitative Analysis of Removed Samples}
\begin{figure*}[!htb]
    \centering
    \includegraphics[width=\linewidth]{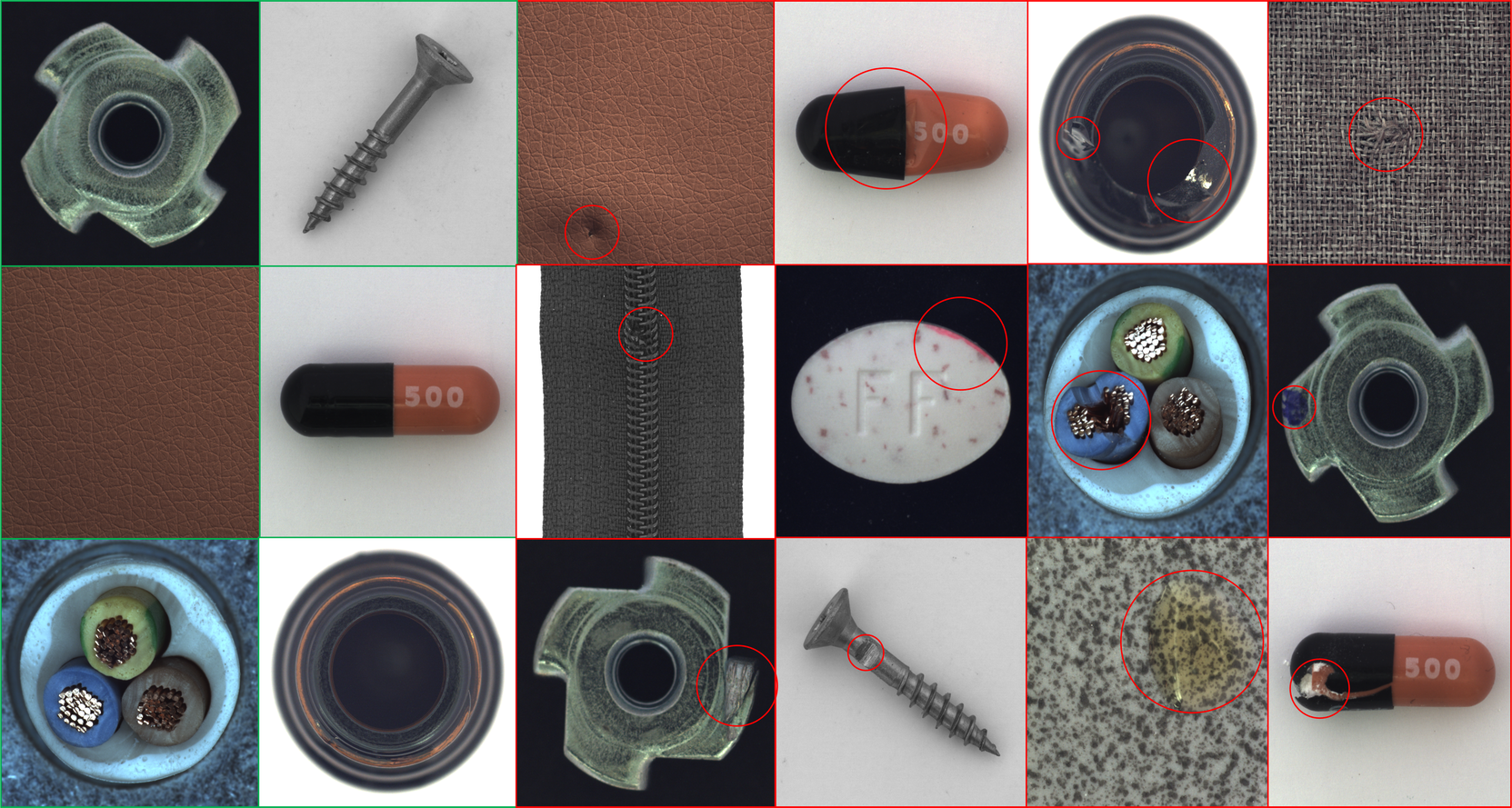}
    \caption{A visual representation of good (outlined in green) and defective (outlined in red) samples under different noise levels. Defects are marked with red circles, illustrating the distribution of flaws within the samples.}
    \label{fig:deleted}
\end{figure*}

 This section presents a detailed quantitative analysis of the samples removed during the defect detection process for both the KSDD2 and MVTec-AD datasets. This analysis is segmented by product categories and noise levels, providing insights into the effectiveness of the Iterative Refinement Process (IRP) in improving defect detection accuracy. Table~\ref{tab2} shows the number of good and bad samples removed at various noise levels (0\% to 50\%) across different product categories within the KSDD2 and MVTec-AD datasets. The categories include Cable, Capsule, Metalnut, Pill, Screw, and Zipper. Each entry in the table specifies the count of good (non-defective) and bad (defective) samples removed during the training and validation phases. The accompanying image Fig.~\ref{fig:deleted} illustrates the samples that were removed during iterative refinement, showcasing both defective samples, which exhibit noticeable defects and noise patterns, and some good samples that were also deleted as part of the refinement process.

The table indicates that at 0\% noise, only good samples are removed. This might seem counterintuitive, but it can be explained by the fact that these good samples are likely close to the decision boundary, making them appear anomalous during the initial iterations of the IRP. Removing these borderline good samples helps refine the model by reducing potential noise and improving its generalization capability. This phenomenon demonstrates the IRP's ability to fine-tune the model even when no bad samples are present. As noise levels increase, the number of bad samples removed rises, which is expected. However, the table also shows that good samples remain removed across all noise levels. This ongoing removal of good samples close to the boundary helps maintain the model's robustness by continuously refining its decision boundaries.

 The quantitative analysis of removed samples underscores the IRP model's capacity to enhance defect detection accuracy through systematic refinement. By removing samples close to the decision boundary, the IRP improves the overall performance, even in the absence of bad samples. As noise levels increase, the model adeptly identifies and eliminates defective samples while maintaining a balance by removing borderline good samples. This approach fine-tunes the model and ensures its robustness and reliability in real-world industrial applications.
 
\section{\uppercase{Discussion}}
The robust performance of the Self-Supervised Iterative Refinement Process (IRP) across varying noise levels underscores its potential utility in industrial settings, where defect patterns may be unpredictable. The model's ability to maintain high AUROC values under increasing noise conditions illustrates its capacity to effectively distinguish subtle defect features from normal variability, which is an essential attribute for critical quality control environments.

While the model demonstrates strong performance, there is scope to enhance its operational efficiency. The current training setup, designed for moderate scale and tailored to specific experimental conditions, serves well under controlled scenarios. However, as we aim to deploy the model in more dynamic industrial settings, optimizing its capacity to handle larger datasets and faster processing cycles will be crucial. These enhancements are intended to reduce the time required to reach model convergence and increase image processing throughput, ensuring the model can meet the demands of real-time defect detection. Future developments could include refining the training algorithms to accelerate convergence without compromising accuracy, and expanding the system's architecture to support more simultaneous operations. These improvements will be critical for deploying the model in environments where rapid decision-making is essential. Furthermore, the detailed analysis of different classes within the MVTec-AD dataset highlights the IRP's adaptability to diverse manufacturing scenarios. The pronounced resilience in classes with intricate defect patterns, such as Screws and Cables, underscores the model's sophisticated feature recognition capabilities, which are likely bolstered by advanced data preprocessing and anomaly scoring mechanisms. Further research could focus on optimizing the model to enhance detection accuracy in scenarios with extremely subtle defect signals, such as those found in high-grade pharmaceutical manufacturing.

Future research efforts might also refine the probabilistic models used within the IRP, enhancing their sensitivity to minor anomalies and incorporating real-time learning capabilities to adapt to new types of defects as they emerge dynamically. Addressing the current limitations by increasing the processing capacity of the model and reducing dependency on lengthy epoch training will be critical. Additionally, exploring the integration of the IRP with other industrial monitoring systems could broaden its applicability, ensuring it remains a versatile and effective tool in the rapidly evolving landscape of manufacturing technologies.

\section{\uppercase{Conclusion}}
\label{sec:conclusion}

The experimental validation of the Self-Supervised Iterative Refinement Process (IRP) presented in this study underscores its substantial efficacy in enhancing anomaly detection across diverse industrial settings. Employing robust statistical measures and sophisticated anomaly scoring mechanisms, the IRP has consistently demonstrated superior performance over traditional methods such as the vanilla model and One Shot Removal (OSR) model, particularly in environments characterized by high noise levels. Through rigorous testing on the KSDD2 and MVTec-AD datasets, the IRP not only achieved high AUROC scores but also exhibited remarkable resilience against variable noise intensities, effectively maintaining its defect detection capabilities even under challenging conditions. The system’s ability to accurately distinguish between defective and non-defective items, even with subtle defect features, highlights its potential as a critical tool for maintaining high standards in industrial quality control. These results validate the effectiveness of the IRP, showcasing its adaptability to different defect types and its robust performance across a spectrum of industrial products. The study’s outcomes suggest that the IRP can greatly enhance the precision and reliability of defect detection systems, ensuring significant improvements in quality assurance processes within manufacturing environments.


\section*{\uppercase{Acknowledgements}}

This study was carried out within the PNRR research activities of the
consortium iNEST (Interconnected North-Est Innovation Ecosystem) funded by the European Union Next-GenerationEU (Piano Nazionale di Ripresa e Resilienza (PNRR) – Missione 4 Componente 2, Investimento 1.5 – D.D. 1058  23/06/2022, \\ECS\_00000043).


\bibliographystyle{apalike}
{\small
\bibliography{example}}

\begin{thebibliography}{}

\bibitem[Akcay et~al., 2019]{akcay2019ganomaly}
Akcay, S., Atapour-Abarghouei, A., and Breckon, T.~P. (2019).
\newblock {GANomaly}: Semi-supervised anomaly detection via adversarial training.
\newblock In {\em Asian Conference on Computer Vision (ACCV)}.

\bibitem[Aqeel et~al., 2024]{aqeel2024delta}
Aqeel, M., Sharifi, S., Cristani, M., and Setti, F. (2024).
\newblock Self-supervised learning for robust surface defect detection.
\newblock In {\em International Conference on Deep Learning Theory and Applications (DELTA)}.

\bibitem[Beggel et~al., 2020]{beggel2020robust}
Beggel, L., Pfeiffer, M., and Bischl, B. (2020).
\newblock Robust anomaly detection in images using adversarial autoencoders.
\newblock In {\em European Conference on Machine Learning and Knowledge Discovery in Databases (ECML-PKDD)}.

\bibitem[Bergmann et~al., 2019]{bergmann2019mvtec}
Bergmann, P., Fauser, M., Sattlegger, D., and Steger, C. (2019).
\newblock Mvtec ad--a comprehensive real-world dataset for unsupervised anomaly detection.
\newblock In {\em Proceedings of the IEEE/CVF conference on computer vision and pattern recognition}, pages 9592--9600.

\bibitem[Bhatt et~al., 2021]{bhatt2021image}
Bhatt, P.~M., Malhan, R.~K., Rajendran, P., Shah, B.~C., Thakar, S., Yoon, Y.~J., and Gupta, S.~K. (2021).
\newblock Image-based surface defect detection using deep learning: A review.
\newblock {\em Journal of Computing and Information Science in Engineering}, 21(4):040801.

\bibitem[Bo{\v{z}}i{\v{c}} et~al., 2021a]{bovzivc2021end}
Bo{\v{z}}i{\v{c}}, J., Tabernik, D., and Sko{\v{c}}aj, D. (2021a).
\newblock End-to-end training of a two-stage neural network for defect detection.
\newblock In {\em International conference on pattern recognition (ICPR)}. IEEE.

\bibitem[Bo{\v{z}}i{\v{c}} et~al., 2021b]{Bozic2021COMIND}
Bo{\v{z}}i{\v{c}}, J., Tabernik, D., and Sko{\v{c}}aj, D. (2021b).
\newblock {Mixed supervision for surface-defect detection: from weakly to fully supervised learning}.
\newblock {\em Computers in Industry}.

\bibitem[Capogrosso et~al., 2024]{capogrosso2024diffusion}
Capogrosso, L., Girella, F., Taioli, F., Chiara, M., Aqeel, M., Fummi, F., Setti, F., Cristani, M., et~al. (2024).
\newblock Diffusion-based image generation for in-distribution data augmentation in surface defect detection.
\newblock In {\em International Joint Conference on Computer Vision, Imaging and Computer Graphics Theory and Applications}, volume~2, pages 409--416. SciTePress.

\bibitem[Chen et~al., 2021]{chen2021surface}
Chen, Y., Ding, Y., Zhao, F., Zhang, E., Wu, Z., and Shao, L. (2021).
\newblock Surface defect detection methods for industrial products: A review.
\newblock {\em Applied Sciences}, 11(16):7657.

\bibitem[De~Vitis et~al., 2020]{de2020row}
De~Vitis, G.~A., Foglia, P., and Prete, C.~A. (2020).
\newblock Row-level algorithm to improve real-time performance of glass tube defect detection in the production phase.
\newblock {\em IET Image Processing}, 14(12):2911--2921.

\bibitem[Defard et~al., 2021]{defard2021padim}
Defard, T., Setkov, A., Loesch, A., and Audigier, R. (2021).
\newblock Padim: a patch distribution modeling framework for anomaly detection and localization.
\newblock In {\em International Conference on Pattern Recognition (ICPR)}.

\bibitem[Girella et~al., 2024]{girella2024cbmi}
Girella, F., Liu, Z., Fummi, F., Setti, F., Cristani, M., and Capogrosso, L. (2024).
\newblock Leveraging latent diffusion models for training-free in-distribution data augmentation for surface defect detection.
\newblock In {\em International Conference on Content-based Multimedia Indexing (CBMI)}.

\bibitem[Hu et~al., 2022]{hu2022robust}
Hu, M., Wang, Y., Feng, X., Zhou, S., Wu, Z., and Qin, Y. (2022).
\newblock Robust anomaly detection for time-series data.

\bibitem[Jawahar et~al., 2023]{jawahar2023leather}
Jawahar, M., Anbarasi, L.~J., and Geetha, S. (2023).
\newblock Vision based leather defect detection: a survey.
\newblock {\em Multimedia Tools and Applications}, 82(1):989--1015.

\bibitem[Luo et~al., 2022]{luo2022robust}
Luo, Z., He, K., and Yu, Z. (2022).
\newblock A robust unsupervised anomaly detection framework.
\newblock {\em Applied Intelligence}, 52(6):6022--6036.

\bibitem[Ono et~al., 2020]{ono2020robust}
Ono, Y., Tsuji, A., Abe, J., Noguchi, H., and Abe, J. (2020).
\newblock Robust detection of surface anomaly using lidar point cloud with intensity.
\newblock {\em The International Archives of the Photogrammetry, Remote Sensing and Spatial Information Sciences}, 43:1129--1136.

\bibitem[Roth et~al., 2022]{roth2022towards}
Roth, K., Pemula, L., Zepeda, J., Sch{\"o}lkopf, B., Brox, T., and Gehler, P. (2022).
\newblock Towards total recall in industrial anomaly detection.
\newblock In {\em IEEE/CVF Conference on Computer Vision and Pattern Recognition (CVPR)}.

\bibitem[Rousseeuw and Hubert, 2018]{rousseeuw2018anomaly}
Rousseeuw, P.~J. and Hubert, M. (2018).
\newblock Anomaly detection by robust statistics.
\newblock {\em Wiley Interdisciplinary Reviews: Data Mining and Knowledge Discovery}, 8(2):e1236.

\bibitem[Rudolph et~al., 2021]{rudolph2021same}
Rudolph, M., Wandt, B., and Rosenhahn, B. (2021).
\newblock Same same but differnet: Semi-supervised defect detection with normalizing flows.
\newblock In {\em IEEE/CVF Winter Conference on Applications of Computer Vision}.

\bibitem[Su et~al., 2019]{su2019robust}
Su, Y., Zhao, Y., Niu, C., Liu, R., Sun, W., and Pei, D. (2019).
\newblock Robust anomaly detection for multivariate time series through stochastic recurrent neural network.
\newblock In {\em ACM SIGKDD International Conference on Knowledge Discovery and Data Mining}.

\bibitem[Tian and Jia, 2022]{tian2022dcc}
Tian, R. and Jia, M. (2022).
\newblock Dcc-centernet: A rapid detection method for steel surface defects.
\newblock {\em Measurement}, 187:110211.

\bibitem[Vrochidou et~al., 2022]{vrochidou2022marble}
Vrochidou, E., Sidiropoulos, G.~K., Ouzounis, A.~G., Lampoglou, A., Tsimperidis, I., Papakostas, G.~A., Sarafis, I.~T., Kalpakis, V., and Stamkos, A. (2022).
\newblock Towards robotic marble resin application: Crack detection on marble using deep learning.
\newblock {\em Electronics}, 11(20).

\bibitem[Zavrtanik et~al., 2021]{zavrtanik2021draem}
Zavrtanik, V., Kristan, M., and Sko{\v{c}}aj, D. (2021).
\newblock Draem-a discriminatively trained reconstruction embedding for surface anomaly detection.
\newblock In {\em IEEE/CVF International Conference on Computer Vision (ICCV)}.

\bibitem[Zavrtanik et~al., 2022]{zavrtanik2022dsr}
Zavrtanik, V., Kristan, M., and Sko{\v{c}}aj, D. (2022).
\newblock {DSR}--a dual subspace re-projection network for surface anomaly detection.
\newblock In {\em European Conference on Computer Vision (ECCV)}.

\bibitem[Zhang et~al., 2022]{zhang2022steel}
Zhang, C., Wang, Z., Liu, B., Xiaolei, W., et~al. (2022).
\newblock Steel plate defect recognition of deep neural network recognition based on space-time constraints.
\newblock {\em Advances in Multimedia}, 2022.

\bibitem[Zhang et~al., 2021]{zhang2021visual}
Zhang, S., Zhang, Q., Gu, J., Su, L., Li, K., and Pecht, M. (2021).
\newblock Visual inspection of steel surface defects based on domain adaptation and adaptive convolutional neural network.
\newblock {\em Mechanical Systems and Signal Processing}, 153:107541.

\bibitem[Zhao et~al., 2019]{zhao2019rad}
Zhao, Z., Birke, R., Han, R., Robu, B., Bouchenak, S., Mokhtar, S.~B., and Chen, L.~Y. (2019).
\newblock Rad: On-line anomaly detection for highly unreliable data.
\newblock {\em arXiv preprint arXiv:1911.04383}.

\bibitem[Zhou and Paffenroth, 2017]{zhou2017anomaly}
Zhou, C. and Paffenroth, R.~C. (2017).
\newblock Anomaly detection with robust deep autoencoders.
\newblock In {\em ACM SIGKDD International Conference on Knowledge Discovery and Data Mining}.

\end{thebibliography}

\end{document}